# Evaluating the Performance of Offensive Linemen in the NFL

Nikhil Byanna, Diego Klabjan

Department of Industrial Engineering & Management Sciences

Northwestern University, Evanston, IL, USA

**ABSTRACT**

How does one objectively measure the performance of an individual offensive lineman in the NFL? The existing literature proposes various measures that rely on subjective assessments of game film, but has yet to develop an objective methodology to evaluate performance. Using a variety of statistics related to an offensive lineman's performance, we develop a framework to objectively analyze the overall performance of an individual offensive lineman and determine specific linemen who are overvalued or undervalued relative to their salary. We identify five players across the 2013-2014 and 2014-2015 NFL seasons that are considered to be overvalued or undervalued and corroborate the results with existing metrics that are based on subjective evaluation. To the best of our knowledge, the techniques set forth in this work have not been utilized in previous works to evaluate the performance of NFL players at any position, including offensive linemen.

# 1. Introduction

The front office of teams in the National Football League (NFL) face critical decisions on a daily basis. From weekly game planning to filling coaching vacancies, teams must constantly evaluate their personnel to make short and long-term decisions that are in the best interest of the franchise. One of the most important decisions that must be made on a yearly, and even weekly, basis is determining which players should be on the 53 man roster.

There are three main avenues through which teams acquire players for their roster: the NFL draft, free agency, and the trading block. To analyze and rank college football players, teams must look at the player's college performance, character, and athleticism. This is very similar to the evaluation of players in free agency and on the trading block, but the main difference is the extent to which a player's past performance can inform a team of their expected future performance. Given the lack of parity in college football, there is often a large amount of variability in the ability of a given team's opponents. Thus, it is difficult to glean the ability of a player solely based on his statistics from college. However, the latter two forms of player acquisition mentioned above – free agency and the trading block— can be more accurately assessed based on a player's past performance in the NFL.

In both free agency and trading, NFL teams must not only assess their personnel needs, but also take into consideration the salary cap, or a league-wide budget constraint capping how much teams can spend on players' salaries each year. In any acquisition of a player, an equilibrium must be reached in contract negotiations in which both parties are satisfied with the agreed upon compensation. A problem that teams face in these contract negotiations involves the notion of asymmetric information. The player has perfect information regarding his ability while teams do not. Thus, the player has leverage in negotiations if the team cannot logically defend their salary offer. Therefore, teams are constantly trying to develop analytical methods to more accurately assess a player's true value, and thus better inform their free agency and trade decisions.

Within the last decade, performance metrics have been developed to evaluate numerous positions in the NFL. However, the offensive lineman position is one of the few positions that has yet to be evaluated objectively on an individual basis given the lack of tangible statistics that are directly related to their performance. According to an interview with the Director of Pro Personnel for an NFL team, the evaluation of offensive linemen is 95% based on watching game film and 5% on using performance metrics and quantitative analysis. Once enough film has been watched for a given player and grades have been assigned to individual games and the lineman as a whole, the scouting team will then find "comparables," or other players who were similar to the lineman in question across various numerical dimensions. These dimensions are typically descriptive statistics such as height, weight, and number of Pro Bowl appearances. Once a team finds 4 to 5 comparables, they determine what would be a reasonable offer to make to the player. While this method has been one of the primary means of evaluation for NFL teams throughout the history of the league, the growing popularity of statistics and analytics in the world of sports has led teams to consider analytics as a method of evaluation in recent years.

The goal of this research is to develop a methodology that can be used to identify offensive linemen that are overvalued and undervalued, thus providing teams with an analytical method that can be used when making salary decisions in free agency and on the trading block. The framework is structured to help an NFL executive answer the following question: does the player in question deserve to be compensated more favorably, similarly, or less favorably than the previous year? The methodology clusters players together based on statistics that are determined to be priced into the salaries of offensive linemen in the NFL labor market, effectively creating a pool of players who deserve to be compensated similarly based on their performance. If a player is found to have a salary that is anomalous as compared to the salaries of other players within his cluster, then he is considered to be either overvalued or undervalued depending on the direction in which his salary is anomalous.

If a player is found to be overvalued, then an NFL executive has reason to believe that the NFL labor market has inefficiently priced that player's salary and the player deserves to be compensated less favorably than in the previous year. Similarly, if a player is found to be undervalued, then an NFL executive has reason to believe that the player is worth more than the salary that he was paid in the previous year. Thus, the question set forth is answered through the analysis proposed.

By creating a framework to analyze a player's performance based on objective statistics from the season in question, we reduce the subjectivity inherent in current metrics used to evaluate offensive linemen, such as the Pro Football Focus metric described in **Section 2**. The novelty of our approach is that it considers the actual outcomes that occur as a result of an offensive lineman's performance on a given play, as opposed to subjectively determining the impact that the given lineman had on the outcome of a play. This allows NFL executives to tie player performance to realized outcomes, which is directly related to the ultimate goal of teams: putting their team in a position to generate outcomes that lead to wins.

In **Section 3,** the data used for the analysis is described. In **Section 4.1**, linear regression techniques are employed to determine the characteristics pertaining to offensive linemen that the NFL labor market prices into player salaries. **Section 4.2** then performs a clustering algorithm to group players together of similar ability based on the characteristics found to be significant in **Section 4.1**, while **Section 4.3** provides characterizations of these clusters based on statistical comparisons. In **Section 4.4, we** fit distributions to the salaries of the players within a cluster to determine the empirical salary distributions of each cluster. **Section 4.5** then uses these distributions to identify players who are both overvalued and undervalued within each cluster. In **Section 5.1** and **Section 5.2**, we discuss our findings and corroborate them with the subjective metrics described above. Finally, **Sections 6** provide a further discussion of the implications of the findings, as well as potential limitations and suggestions for future research, while **Section 7** concludes.

## 2. Related Work

The position in the NFL that has been the easiest to analyze is the quarterback. The quarterback not only touches the ball on every play that his team is on offense, but the statistics associated with a quarterback are very easy to track. In 1973, Don Smith of the Pro Football Hall of Fame and Seymour Siwoff of Elias Sports Bureau created the quarterback rating metric, one of the most widely cited statistics in the NFL (*NFL's Passer Rating*). The idea behind the quarterback rating was to create a standardized metric that could be used to compare performances from game to game and season to season.

The quarterback rating was recently expanded on when ESPN unveiled the metric "Total Quarterback Rating (Total QBR)," which attempts to measure the situational performance, as opposed to the absolute performance, of the quarterback (Oliver, 2011). The ability to contextualize a quarterback's performance and determine the quarterback's exact contribution to an outcome makes Total QBR a widely accepted and used metric in the NFL. A large component of Total QBR is the concept of Win Probability Added (WPA), which measures the impact of a given play on the team's chances of winning the game (Burke, 2010). Total QBR attempts to measure the quarterback's contribution to the WPA of the team over the course of the game. That is, for any given offensive play, Total QBR captures the exact impact that the quarterback's decisions had on the win probability of the team. While Total QBR utilizes the concept of WPA, it also takes into consideration the Expected Points Added (EPA) of each play, which is similar to WPA but measures the impact of a given play on the team's expected total points for the game. By using a combination of these two metrics, Total QBR creates a robust metric to measure a quarterback's performance. If a universal metric like Total QBR could be created for other positions in the NFL, teams would have a much easier time comparing players based on their statistical performance.

The concept of WPA and EPA has been applied to the majority of positions in the NFL by Advanced Football Analytics, a website dedicated to the analysis of the NFL using mathematical and statistical models. The two metrics have only been calculated independently however, and thus there has been no universal metric such as Total QBR created for any position other than quarterback. Furthermore, the only position that has not had WPA and EPA applied to individuals is the offensive lineman: left and right guard, left and right tackle, and center. Advanced Football Analytics currently has a WPA and EPA calculated for the offensive line as a whole, but has no WPA or EPA calculation for an individual lineman. Given the lack of tangible statistics for offensive linemen, the majority of football analysts have had the same difficulty that Advanced Football Analytics has had in evaluating the performance of an individual offensive lineman.

The most widely cited and used metric for individual offensive lineman is published by Pro Football Focus (PFF), a website published by PFF Analysis Ltd. PFF grades every single snap on offense, defense, and special teams, assigning a grade to each player on the field. These grades are obtained by watching film from every game in an NFL season. The grades range from +2.0 to -2.0 in increments of 0.5, with a grade of "0" being viewed as the "expected" grade for an NFL player. The main factors considered for an offensive lineman are pass protection, run blocking, screen blocking, discipline, and procedure (see reference *Grading*). These grades are aggregated across the season and normalized across the position before a final rating is produced. This methodology appears to be sound and is the only notable metric that has been published thus far in relation to offensive linemen. However, one major aspect of the methodology is the subjectivity involved in grading each play. While some of the subjectivity can be controlled for by using consistent graders and validating across graders, the nature of the system is inherently subjective. Based on current research, there has yet to be an objective methodology created to evaluate the performance of an individual offensive lineman, which is what this research aims to develop.

# 3. Data

The primary dataset used for analysis was obtained from STATS LLC, a global statistics and sports information company that tracks, analyzes, and distributes play-by-play data from a variety of sports. The data contained game-by-game data for every offensive lineman that was recorded as having played at least 1 snap in a regular season or playoff game from the 2013-2014 and 2014-2015 seasons. In total, there were 5383 data points with 44 variables for each point. The variables are detailed in **Table 1**:

| Descriptive Variables | Other |
| --- | --- |
| Unique game code | Base salary |
| Playoff vs Regular Game | Signing bonus |
| Game date | Incentives |
| Player name | Cap Value |
| Team | Snaps |
| Opponent | Holding penalties on rush attempts |
| Position | Holding penalties on pass attempts |
| Rookie year | |
| Draft round | |
| Draft pick | |
| Birthday | |

| Rushing Statistics | Passing Statistics |
| --- | --- |
| Rush attempts to side | Passing yards |
| Rush attempts not to side | Dropbacks |
| Stuffs to side | Passing attempts |
| Stuffs not to side | Passing completions |
| Rush yards to side | Sacks to side |
| Rush yards not to side | Sacks not to side |
| Yards after contact to side | Sack yards to side |
| Yards after contact not to side | Sack yards not to side |
| Rush touchdowns to side | Pressures to side |
| Rush touchdowns not to side | Pressures not to side |
| Successful rushes to side | Hurries to side |
| Successful rushes not to side | Hurries not to side |
| | Knockdowns to side |
| | Knockdowns not to side |
| | Quarterback release time |
| | Quarterback release attempts |
| | Release time under pressure |
| | Release attempts under pressure |

**Table 1:** List of variables contained in the dataset

As seen in **Table 1**, there are statistics that cover a wide range of a player's attributes, including attributes related to the game, attributes related to the player's salary, and attributes related to the running and passing plays in which the player was involved. Each statistic was tracked on a play-by-play basis such that it was only included for a given player if the player was on the field during the play. The play-by-play statistics were then aggregated on a game-by-game basis. Furthermore, the majority of statistics include a "to side" and "not to side" entry. These statistics were determined based on the location in which the play occurred. Based on the tracking methodology of STATS, the line is divided into five distinct splits –LS, L, M, R, RS— where LS indicates the far left side of the line, RS indicates the far right side of the line, and the remaining 3 categories are evenly divided between LS and RS. For a given position, the "to side" statistics were aggregated based on any play that was recorded as corresponding to a split in column 2 of **Table 2**. For that same position, the "not to side" statistics were then aggregated based on any play that was recorder as going to any split not listed in column 2 of **Table 2**:

| Position | Splits that are "to side" |
|----------|---------------------------|
| LT | LS, L |
| LG | LS, L, M |
| C | L, M, R |
| RG | M, R, RS |
| RT | R, RS |

**Table 2:** Splits of the line that are categorized as "to side" for a given position

The rationale behind this methodology is as follows. The reason that it has been difficult to objectively evaluate an offensive lineman's performance is because it is nearly impossible to track individual statistics for linemen. One way to approach this impediment is to track statistics that the offensive lineman is responsible for ("to side"), as well as the statistics that the offensive lineman is not responsible for ("not to side"). The "to side" statistics provide information on the direct contribution that the offensive lineman makes on a given play. The "not to side" statistics effectively serve as a control, showing how the statistics differ based on the other players on the field. For example, if an offensive lineman on a given team has a large number of "to side" rushing yards but also has a large number of "not to side" rushing yards, this indicates that the numbers might be influenced by an exceptional running back or exceptional offensive lineman on the other side of the line. Based on this logic, "differential statistics" were created to highlight the differential between the "to side" and "not to side" statistics. In most cases, a positive differential indicates that the player outperforms the linemen on the other side, while a negative differential indicates that the linemen on the other side of the line outperform the player in question. A derived attribute was created for every statistic in **Table 1** that includes a "to side" and "not to side" entry.

Offensive linemen are often involved in run plays that are directed to the other side of the line – most notably in zone or power schemes. Ideally, a dataset used for this analysis would include a variable that indicates whether a specific lineman was involved in a block when the ball is run to the other side of the line, in the instance of a stretch, pull, or counter play. In this instance, our methodology could be adapted to calculate the differential between plays that involved the offensive lineman and plays that did not involve the offensive lineman in such plays, which would serve as a control for the ability of other players on the field. However, this type of data is currently not tracked, which leads to the methodology described above as the best proxy to control for the ability of other players on the field. A further discussion of the use of "differential statistics" can be found in **Appendix A**.

Aside from the data acquired from STATS, additional data needed to be extracted from the internet to have a full dataset that captures all aspects of an offensive lineman's performance. The STATS dataset was able to provide substantial information regarding an offensive lineman's performance on a game-by-game basis. However, a key measure of an offensive lineman's overall performance that could not be pulled from the STATS database is the player's past performance, which can be represented by the number of selections the player has had to the Pro Bowl and All-Pro teams. While the number of Pro Bowl selections is an imperfect measure of a player's ability, due to players receiving votes simply because they were highly touted coming out of college, the Pro Bowl is generally regarded as displaying the best players in the NFL, and thus the number of Pro Bowl selections can be used as a proxy for a player's ability. In contrast to the Pro Bowl, the All-Pro teams are selected by renowned sports writers and columnists and are a much more accurate representation of a player's performance being considered as exceptional. There are 3 different All-Pro teams that are selected: Associated Press 1$^{st}$ Team, Associated Press 2$^{nd}$ Team, and Pro Football Writers 1$^{st}$ Team, in order of respective prestige. Thus, there were four additional variables created in the data set, one for the number of Pro Bowl selections and one for the number of selections to each of the three All-Pro teams. The data was manually pulled from Pro-Football Reference and merged with the existing STATS dataset (Pro Football Reference, 2014).

The last data set that has been gathered involved the Pro Football Focus grades described in **Section 2**. The individual grades for each offensive lineman were extracted from 2007 – 2015, the years that the data is available. These grades are used for two purposes. The first purpose is as potential independent variables in the regressions outlined in **Section 4.1**, and the second purpose is to serve as a validation mechanism for the findings in our work. The data has been manually extracted from the Pro Football Focus website and merged with the existing STATS dataset (Pro Football Focus, 2015).

# 4. Methodology
## 4.1 NFL Labor Market Pricing of Offensive Lineman

In order to appropriately identify overvalued and undervalued offensive linemen, it must first be determined how the NFL labor market prices the salaries of offensive linemen. The first analysis seeks to determine which player characteristics, both performance and experience based, are valued in the NFL labor market. Using a player's average salary per year over the entirety of the contract (subsequently denoted as Cap Value), adjusted for cap inflation, as the dependent variable in a linear regression, we can determine which characteristics are priced into a player's salary by examining the independent variables that are statistically significant in the model specification. The model specification reads:

$$Cap\ Value_i = \mathbf{X}_i \cdot \boldsymbol{\alpha} \qquad (1)$$

In this model specification, **X** represents the vector of values corresponding to the independent variables in the model for player *i*, including a 1 as the first entry to account for the intercept of the model. $\boldsymbol{\alpha}$ represents a column vector of coefficients corresponding to the independent variables, with the first entry of the vector corresponding to the intercept. A preliminary stepwise regression can be run using the data to determine a linear model specification that accurately characterizes the data. The initial set of potential variables includes 15 variables corresponding to the data described in **Section 3.** This includes demographic characteristics (e.g. age, experience, draft round and pick), proxies for past performance (e.g. All Pro and Pro Bowl selections, average of past season PFF +/- ratings) and 5 "differential statistics" for the year in question believed to be important in describing a lineman's performance (e.g. stuff percentage differential, rushing yards per carry, successful run percentage, pressures allowed percentage, sacks allowed percentage).[1] The player's PFF rating for the year in question was also included as a potential predictor, to assess whether it is a stronger representation of the player's on field performance in a given year than the "differential statistics" used.[2]

It is worth noting that certain players were excluded from the dataset before the regression model was run. Any player in the dataset who was still under their rookie contract was eliminated from the dataset and excluded from the remainder of the dataset. The reason for this is because rookie contracts are determined based on a fixed scale and thus are not representative of a free market in which the true value of a player can be determined. Additionally, any players who were not unrestricted free agents at the time of signing their existing contract were excluded. This is due to the fact that the team they were with at the time of contract negotiation owned additional rights to the player that they otherwise would not have if the player were an unrestricted free agent.[3] Lastly, only players who signed contracts since 2011 (when the new CBA was signed) were included in the regression data.

---

[1] See Appendix B for a more detailed description of the differential statistics
[2] It is worth noting that the player's PFF rating for the current year was not found to be significant in the regression, indicating that the differential statistics that are used have a stronger correlation to a player's salary than PFF ratings.

The initial regression model serves to preliminarily explore the significance of various independent variables. While this model considers a variety of factors specific to the offensive lineman in question, a factor that it omits is the ability of the other offensive linemen on his line. As described earlier, the derivation of differential statistics is able to serve as a control for the ability of linemen on the opposite side of the line, but there is currently no control in the model for the ability of linemen on the same side of the line as the lineman in question. Based on this omission, a control was created to account for the ability of the linemen on the same side of the line, to determine whether this is significant in a regression that models an offensive lineman's salary.

To create this metric, the individual predictors from the initial regression model are categorized into "experience" predictors and "performance" predictors. "Performance" predictors are any predictors that directly relate to the player's performance on the field during the season. All other individual predictors are categorized as "experience" predictors. We denote by $E$, $P$ the set of all "experience, performance" predictors, respectively. The rationale behind categorizing the predictors into these two sets is so that two distinct metrics can be created: one that captures the performance of a player and one that captures the experience of a player. To calculate these metrics, we seek to assign a relative weight to each predictor based on its effect on a player's salary, which is represented by the coefficient of the predictor that results from the regression model. Thus, using the coefficients of each predictor, $\alpha_j$, and the set that the predictor belongs to ($E$ or $P$), weights can be calculated for each of the predictors of the model and can subsequently be used to calculate the desired metrics. These normalized weights, denoted as $\gamma_j$, are constructed as follows:

$$\gamma_j = \frac{\alpha_j}{\sum_{i \in E} \alpha_i} \quad if\ j \in E \tag{2}$$

$$\gamma_j = \frac{\alpha_j}{\sum_{i \in P} \alpha_i} \quad if\ j \in P \tag{3}$$

---

[3] Players with two separate observations (one for the 2013-2014 season and one for the 2014-2015 season) that were under the same contract had only one data point included in the regression data due to heteroscedasticity concerns. Their statistics from the two seasons were averaged to create a single data point. After the regression model was determined, these two observations were then disaggregated for the purposes of clustering analysis.

These weights allow an overall "Experience" and "Performance" metric to be created for each player $i$ in the dataset:

$$PerformanceMetric_i = \sum_{j \in P} \gamma_j \cdot X_{i,j} \quad \forall i \in \{1, \ldots, n\} \tag{4}$$
$$ExperienceMetric_i = \sum_{j \in E} \gamma_j \cdot X_{i,j} \quad \forall i \in \{1, \ldots, n\} \tag{5}$$

By creating these two metrics, we can include controls for the abilities of other players on the same side of the offensive line. For each player in the data set, the Experience metric for the player that played on the left and right side of the player in the given season are averaged to get a "Team Experience Metric." Likewise, the Performance metric for the player that played on the left and right side of the player in the given season are averaged to get a "Team Performance Metric." It is worth noting that players who play Left and Right Tackle, which means that they are at the end of the line, only have one player playing next to them, and thus had the team metrics take on the value of the metric for the one player that played next to them. These new statistics are now included as potential independent variables in a new stepwise regression model, along with all of the original independent variables, to create a final linear regression model that is used in the remainder of the analysis.

## 4.2 Clustering

The linear regression model, formulated after performing the stepwise regression from the previous section with $m$ independent variables, determines which player characteristics are valued by the NFL labor market, and thus priced into the salaries of offensive linemen. We next seek to group similar players together based on a comparison of the characteristics specified in the model. This is done via a k-means cluster analysis, which seeks to create *k* distinct clusters of players from the overall dataset of *n* players (Hartigan & Wong, 1979), who each containing *m* standardized attributes.

To determine the optimal number of clusters, the k-means clustering algorithm is run for values of $k \in \{1, ..., 20\}$ and the Krzanowski-Lai statistic (Krzanowski & Lai, 1988) is computed for each iteration of *k*, as defined by:

$$C_k = \left|\frac{DIFF(k)}{DIFF(k+1)}\right| \text{ where } DIFF(k) = (k-1)^{\frac{2}{m}} \cdot Within\ SS_{k-1} - k^{\frac{2}{m}} \cdot Within\ SS_k \qquad (6)$$

Here "*Within SS*" is the sum of square distances within all clusters. Once $C_k$ is determined for $k \in \{1, ..., 20\}$, it can then be plotted as a function of *k* to determine the optimal number of clusters. To determine $k^*$, we identify all *k* values that correspond to local maxima of $C_k$ as potential candidates and then further examine the data to choose from these candidates. After $k^*$ is determined, the k-means clustering algorithm is run with $k^*$ clusters.[4]

## 4.3 Characterization of Clusters

The goal of forming the $k^*$ clusters created in the previous section is to group players of similar ability into the same cluster, thereby providing a basis for player comparison. Based on the objective function of the clustering algorithm, a player will theoretically be placed in a cluster such that the distance from him to all other players in the cluster is very small. Thus, a player should be placed in a cluster with other players who are similar to him across the *m* dimensions in the input matrix **X**. The end goal of the analysis is to identify players within the clusters that are overvalued and undervalued relative to their performance. However, the clusters must first be inspected in an attempt to provide a characterization based on the attributes of the players within the cluster.

To create a consistent method of inspection across clusters, t-tests are performed using the sample mean of each predictor within a given cluster and the sample mean of the overall dataset. The t-test is carried out under the null hypothesis that the population mean of the cluster is equal to the overall population mean, with the alternative hypothesis being that the population mean of the cluster is not equal to the overall population mean.

---

[4] Refer to **Appendix C** for a detailed analysis of the k-means application.

This hypothesis test is carried out for each predictor *j* within each cluster *w*, with p-values calculated for each hypothesis test. Using a significance level of 1%, predictors within each cluster that had p-values that are less than .01 are deemed to be statistically significant. Thus, there is evidence that suggests that the population mean of cluster *w* is different from the population mean of the overall population for any predictor *j* that has a corresponding p-value of less than .01. Once these predictors are identified for each cluster *w*, a qualitative assessment of the p-values is conducted to determine whether subsequent hypothesis testing (described in **Section 4.4** and **Section 4.5**) for overvalued and undervalued players should be one or two sided. If the predictors with significant p-values all suggest that a player should be either compensated with a high or low salary, then a one-sided test should be conducted. For example, if a cluster is found to have less experience and below average run blocking performance in the given year, then the cluster should theoretically be compensated with low salaries. Thus, we would want to test for players within that cluster that are overvalued. In other words, we want to find players who are paid a significantly higher salary than the rest of the players in the cluster even though they are not worthy of a higher salary. If a cluster does not contain clear indications such as the preceding example, then a two-sided test should be performed on the cluster to identify both undervalued and overvalued players.

## 4.4 Distribution Fitting

As mentioned in the previous section, the final component of the analysis is to test for overvalued and undervalued players within the clusters that were created. This test is based on the underlying assumption that the empirical salary data is randomly drawn from some parametric distribution for each cluster *w*. Therefore, these empirical salary distributions must be determined before proceeding to test for overvalued and undervalued players. To estimate the parameters for the empirical salary distribution of each cluster, statistical software is used to fit a distribution to each cluster based on the salary values associated with each player within the cluster. The empirical distribution of salaries for any cluster is bounded below by the minimum base salary of an NFL contract, which has been historically increasing on a yearly basis. Thus, for the purposes of distribution fitting, the lower bound of a potential distribution should be the minimum base salary across all years included in the data set.

Based on this information, a select group of empirical distributions are plausible to model the salary of a given cluster. Furthermore, the Lognormal, Gamma, Beta, Pareto, and Weibull distributions have been empirically found to be descriptive models for the distribution of income (McDonald, 1984). Thus, the distributions used for consideration should be restricted to these 5 families of distributions insofar as they can provide a reasonable fit based on the criteria described in the remainder of this section. The first criteria on which to evaluate the fit of a distribution are the Chi-Squared statistic and the Akaike information criteria, which are both statistical measures of goodness of fit. Evaluating these values for a given distribution, in relation to the other candidate distributions, provides a proxy for the relative fit of the distribution. Once the list of candidate distributions is narrowed down based on these statistics, an examination of the P-P (probability-probability) and Q-Q (quantile-quantile) plots will help to inform which distribution to choose. Based on these two criteria, a distribution is chosen for each of the clusters created in **Section 4.3**. For clusters of size $n \leq 15$, it is recommended that distributions are not fit to the data given the small sample size, and thus such small clusters are neglected from further analysis.[5]

## 4.5 Player Identification

---

[5] Refer to **Appendix D** for sample distribution fits and P-P plots for the analysis conducted.

The final component of the analysis, testing for overvalued and undervalued players, can now be carried out using the distributions determined in the previous section. When an NFL executive needs to determine the salary that she or he wants to offer a player, s/he is able to choose from three distinct alternatives: offering the player more money than he made in his previous contract, offering the player the same amount of money, or offering the player less money. All three of these alternatives require a logical argument that the executive must make to the player as to why he warrants the salary that is being offered.

The goal of this analysis, in the case of overvalued players, is to provide executives with a statistical result that shows that the player in question does not deserve to be compensated in a similar fashion to how they were in their previous contract. This result can be formalized by utilizing the distributions found in **Section 4.4**. Consider a highly compensated player $i$ in cluster $w$ with a cumulative distribution function $f$. Suppose an NFL executive chooses a player from cluster $w$ and believes that his performance in the following year will be similar to his past year's performance. Based on this supposition, the input vector $x_i$ would be constructed such that player $i$'s performance from the following season would be expected to be placed in cluster $w$. The question that an NFL executive must ask is if s/he wants to pay the player a salary that is greater than or equal to his previous contract's salary $S$. The probability that a player that is placed in cluster $w$ is paid greater than $S$ dollars is equal to $P(x \geq S) = 1 - f(S)$. If $P(x \geq S) \leq .05$, then $S$ is significant at the 5% level, and an NFL executive has reason to believe that the player should have a compensation lower than $S$ in the coming year. Similar logic is applicable to lower paid players.

Based on the intuition outlined above, the analysis can be carried out for each individual cluster while performing one sided or two sided tests based on the characterizations from **Section 4.4**. From these tests, a list of overvalued and undervalued players is created as the output of the analysis. However, one additional assumption must be verified to create the final list of players that are determined to be overvalued or undervalued. An underlying assumption that is needed to apply the methodology's logic is that each cluster is relatively homogeneous. In other words, the characteristics of the players within a cluster are close enough in Euclidean length that one can consider any two players within a cluster to be similar. This is an assumption that, when violated, weakens the argument set forth in the remainder of the analysis. This concern arises in the analysis conducted given the small sample size. A way to formalize the concept of cluster homogeneity is through the silhouette value for a given point within a cluster analysis (Rousseeuw, 1987). Let $s(i)$ be the resulting silhouette value of player $i$. If $s(i)$ is close to one, player $i$ is very well clustered. If $s(i)$ is close to 0, player $i$ is close to being placed in a neighboring cluster. If $s(i)$ is close to -1, player $i$ would be more appropriately clustered if placed in its neighboring cluster.

Given the construction of $s(i)$, we seek an average $s(i)$ value in the sample that is positive and significantly larger than 0 to be consistent with the assumption of cluster homogeneity. Given the small sample used for this analysis, the average $s(i)$ value was lower than desired, taking on a value of slightly greater than 0.16. Theoretically, as more seasons of data are added to the sample, the clustering algorithm will be able to create more homogenous clusters and the $s(i)$ value will increase to a value that is sufficiently large to be consistent with the assumption of cluster homogeneity (specific thresholds have not been empirically defined, but *Kaufman and Rousseeuw* suggest a value at least above .25 and ideally above .5). To account for the low average $s(i)$ value in the sample, player $i$ has to satisfy the criteria outlined previously in this section, as well as the criterion outlined in (7), to be considered overvalued or undervalued in the analysis:

$$s(i) \geq \frac{\sum_{j=1}^{n} s(j)}{n} \qquad (7)$$

The criterion in (7) stipulates that the silhouette value of player $i$ must be greater than the average silhouette value of the entire sample. It is worth noting that this is a simple heuristic that was used for the purpose of the analysis, and that other heuristics can be tested to develop a stronger condition that must be satisfied.

# 5. Results

## 5.1 Player Clusters

The analysis was conducted using the dataset outlined in **Section 3** and the methodology set forth in **Section 4**. The clustering algorithm was run using 7 clusters[6], with the descriptive statistics of each of the 7 clusters, as well as the entire sample, shown in **Table 3**.

| Cluster | # Players | Mean | Standard Deviation | Median | Minimum | Maximum |
|---|---|---|---|---|---|---|
| 1 | 25 | $ 5,118,847.06 | $ 2,386,079.12 | $ 6,067,019.30 | $ 528,701.62 | $ 9,588,292.68 |
| 2 | 17 | $ 2,967,692.92 | $ 1,834,427.10 | $ 2,984,758.11 | $ 541,463.41 | $ 5,714,855.64 |
| 3 | 18 | $ 3,838,068.63 | $ 1,817,446.69 | $ 4,004,880.79 | $ 795,634.99 | $ 6,731,317.82 |
| 4 | 23 | $ 2,206,081.26 | $ 1,236,261.33 | $ 1,759,045.86 | $ 666,587.32 | $ 5,082,092.17 |
| 5 | 24 | $ 2,528,372.25 | $ 1,288,775.69 | $ 2,523,431.23 | $ 590,243.90 | $ 5,283,149.56 |
| 6 | 8 | $ 6,656,008.67 | $ 1,215,143.94 | $ 6,258,929.77 | $ 5,719,508.25 | $ 9,253,297.36 |
| 7 | 18 | $ 3,098,081.01 | $ 1,925,532.67 | $ 2,559,429.46 | $ 528,701.62 | $ 6,889,518.41 |
| Sample | 133 | $ 3,518,357.31 | $ 2,146,048.26 | $ 3,085,988.93 | $ 528,701.62 | $ 9,588,292.68 |

**Table 3**: Descriptive statistics of salary for each of the 7 clusters and the entire sample

Based on the sizes of the clusters, it was determined that Cluster 6 was not sufficiently large to proceed with the remainder of the analysis. Thus, Clusters 1, 2, 3, 4, 5, and 7 were further analyzed in order to provide a qualitative characterization and determine whether to test the cluster for undervalued players, overvalued players, or both. **Table 4** depicts the characterization of each cluster, based on the analysis of the p-values that resulted from the t-test methodology described in **Section 4.3**.

---

[6] See **Appendix B** for an explanation of why $k^*$ was chosen to be 7

| Cluster | Characterization | Value to test for |
|---|---|---|
| Cluster 1 | Players who were early draft selections and have above average run blocking abilities | Undervalued |
| Cluster 2 | Players who had above average PFF ratings prior to their contract being signed | Undervalued |
| Cluster 3 | Players who were early draft selections, have less experience, and fewer pro bowl selections | Both |
| Cluster 4 | Players who were late draft selections, have below average PFF ratings prior to contract being signed | Overvalued |
| Cluster 5 | Players who were late draft selections, have fewer Pro Bowl selections and below average run blocking abilities | Overvalued |
| Cluster 7 | Players who had below average PFF ratings prior to contract being signed | Overvalued |

**Table 4**: Characterization of Clusters 1, 2, 3, 4, 5, and 7

## 5.2 Player Identification

Employing the methodology described in **Section 4.5** to clusters 1, 2, 3, 4, 5, and 7, five players were identified as being overvalued or undervalued in the 2013-2014 and 2014-2015 seasons combined. It is worth noting that the original analysis identified eleven overvalued and undervalued players, but the additional criterion in (7) eliminated six of the twelve players. The two undervalued players can be found in **Table 5**, along with the given year, the team they were on, the position they played, and their cap value.

| Player | Year | Team | Position | Cap Value |
|---|---|---|---|---|
| John Jerry | 2014 | New York Giants | G | $ 795,635 |
| Mike McGlynn | 2014 | Kansas City Chiefs | G | $ 1,037,594 |

**Table 5:** Players found to be undervalued based on the analysis

These findings suggest that both of these players, based on their 2013 or 2014 characteristics, were similar to players in the dataset who were paid a significantly higher salary. Based on this conclusion, it would be in the best interest of executives to offer these players contracts that provide them with higher compensation than they were given in 2013 or 2014. That being said, there are many other considerations an executive must contemplate given the situation they are facing, which will be discussed further in **Section 6.1**. The analysis also identified three overvalued players, which can be found in **Table 6**, along with the given year, the team they were on, the position they played, and their cap value. For a detailed explanation of how the methodology outlined in **Section 4.1** –

**Section 4.5** was applied to the dataset and produced the findings outlined in this section, please refer to **Appendices B, C, and D**.

| Player | Year | Team | Position | Cap Value |
|---|---|---|---|---|
| Scott Wells | 2013 | St. Louis Rams | C | $ 5,283,150 |
| Scott Wells | 2014 | St. Louis Rams | C | $ 5,283,150 |
| Davin Joseph | 2013 | Tampa Bay Buccaneers | G | $ 6,889,518 |

**Table 6:** Players found to be overvalued based on the analysis

When inspecting the statistics of the 3 overvalued players identified in **Table 6**, the player's performance in the season identified was significantly worse than the average performance of all players in the sample. Davin Joseph and Scott Wells both had significantly lower yards per attempt differentials than the league average, as well as significantly worse stuff percentage differentials. Davin Joseph also had a significantly lower average PFF rating for his time in the league prior to the contract being signed.

These statistics help strengthen the findings, showing that the three players identified as being overvalued did indeed perform significantly worse during the season than other linemen in the league. Furthermore, the actions taken by these players' teams following the season help to further corroborate the findings. Davin Joseph was released by the Buccaneers following the 2013 season, and Scott Wells was released by the Rams following the 2014 season.

## 5.3 Comparison to Pro Football Focus Metrics

Given the lack of published metrics that evaluate individual offensive lineman, it is difficult to use multiple methods to validate the findings in an empirical manner. As mentioned in **Section 2**, the most widely cited metric used to evaluate individual offensive lineman is the Pro Football Focus rankings that are published on a weekly basis. A way to judge whether a player over performed or underperformed relative to their salary is to compare their relative performance ranking, as judged by Pro Football Focus, with their relative salary ranking based on each player's cap value from the given season. If a player's relative performance ranking is lower than their relative salary ranking, then they over performed relative to their salary.[7] Conversely, if a player's relative performance ranking is higher than their relative salary ranking, then they underperformed relative to their salary. This intuition provides a basis for evaluation of whether a player is overvalued or undervalued solely based on their performance from the given season, which will be used to validate the analysis. **Table 7** highlights the relative performance and salary ranking of each of the five players found to be overvalued or undervalued, as compared to every other player in the sample that plays the same position as the player in question.[8]

| Player | Year | Position | Relative Performance Rank | Relative Salary Rank |
|---|---|---|---|---|
| John Jerry | 2014 | G | 23 | 28 |
| Mike McGlynn | 2014 | G | 31 | 27 |
| Scott Wells | 2013 | C | 11 | 3 |
| Scott Wells | 2014 | C | 13 | 3 |
| Davin Joseph | 2013 | G | 30 | 1 |

**Table 7**: Comparison of relative performance rank and relative salary rank for identified players

As seen in **Table 7**, the comparison of relative performance rank to relative salary rank corroborates the findings from this paper for four of the five players identified. In the case of one of the two players found to be undervalued, their relative performance rank was significantly lower than their relative salary rank, suggesting that they over performed relative to their salary. In the case of all three players found to be overvalued, their relative performance rank was significantly higher than their relative salary rank, suggesting that they underperformed relative to their salary. These findings suggest that the methodology presented in this paper may be able to more strongly identify players who are overvalued rather than undervalued, while also looking at the larger differential between relative performance and salary rank for overvalued players versus undervalued players. The discussion in **Section 5.2** regarding overvalued players also supports this hypothesis. However, further application of the methodology to larger datasets would need to be conducted to validate these findings. Furthermore, the findings suggest that this analytical approach should not be used as the only method of player evaluation, but rather in conjunction with other evaluation methods to help defend salary decisions.

---

[7] The relative performance ranking and relative salary ranking are both computed in ascending order. Thus, the player with the best performance ranking has a relative performance ranking of 1, while the player that is paid the highest salary has a relative salary ranking of 1.

[8] For the purposes of this validation, the three positions considered were center, guard, and tackle. Thus, right guards and left guards were grouped together, as well as right tackles and left tackles.

Why do NFL teams not just use a comparison of relative performance rank and relative salary rank on a yearly basis to determine players that are overvalued and undervalued? The main reason we believe this is not a sound approach for evaluation is the subjectivity inherent in the formulation of the performance ranks. The performance rank is based on judgment, whereas the analysis in this work is data driven. Furthermore, we argue that having a relative performance rank that is significantly higher or lower than your relative salary rank is a necessary but not sufficient condition to be considered overvalued or undervalued. If a player is identified as having a significantly higher or lower performance ranking than his relative salary ranking, this is solely indicative that the player's performance in the given year is not worthy of the salary that he has been paid. However, this assessment does not consider the past performance of the player, which is captured in statistics such as experience, age, and awards that the player has received. A player with multiple All-Pro $1^{st}$-team selections that underperforms in a given year relative to his salary may not be considered overvalued if the season was an anomaly. The clustering algorithm used in this analysis accounts for these past performance indicators and determines whether they are strong enough to outweigh a poor performance in a given season by a player. Based on this intuition, the analysis outlined in this paper provides a result that is stronger evidence to determine if a player is overvalued or undervalued.

# 6. Discussion
## 6.1 Considerations for NFL Executives

**Section 5.2** presents an objective result that can inform NFL executives of players that can be considered to be overvalued or undervalued relative to their performance. However, this does not suggest that a team should with absolute certainty give a player a significantly more lucrative contract if that player was found to be undervalued. Likewise, it does not suggest that a team should with absolute certainty release or trade away a player that was found to be overvalued. Consider a player who is found to be undervalued based on this methodology. Although this informs an executive that the player's overall season performance warrants a salary increase, it does not provide information regarding the trend of the player's performance over the course of the season, among other underlying factors not described by the analysis. Thus, once players are identified via the methodology presented in the paper, NFL executives are encouraged to further evaluate the players to ensure that there are no in-season trends that suggest that evaluating a player based on his average statistics over the course of the season would not be an accurate assessment of the player's performance.

Along with further evaluating the player, an NFL executive must also consider the entire salary structure of the team, as well as the organizational structure, before making a definitive decision related to the type of contract s/he wants to offer a player. While this analysis provides a conclusive recommendation as to whether a player should be offered more or less money, it does not provide executives with a player's true value, or in other words the exact salary that a player deserves.[9] Thus, it is imperative for an executive to consider the marginal increase or decrease in performance that would result from signing or releasing an identified player, while also considering the marginal increase or decrease in cost that would result. There exists no universal equilibrium when considering this trade off given the different states that a team may be in regarding their salary cap space and the utility that they derive from having a higher performing team. Therefore, this trade off must be evaluated on a situational basis.

---

[9] Using the regression methodology outlined in **Section 4.1**, it is possible to create a linear model with salary as the dependent variable that can be used to predict a player's salary, or true value. However, given the limitations of the small sample, the predictive power of the model was not considered to be strong enough to include in the analysis.

This analysis is most helpful for teams that have minimal salary cap space and must make financial decisions that will allow them to be comfortably under the cap. In the case of undervalued players, the team can potentially exploit the market inefficiency and sign a player for less than he is worth, while still acquiring a player who performed relatively well in the prior season. In the case of overvalued players, the team can attempt to trade the player to free up salary cap space that they can use to sign a different lineman or use to address other positional needs. Once again, this decision should be made in the best interest of the team's performance as well, which introduces other factors into the decision that cannot be informed by this analysis. This analysis is helpful in providing teams with candidates to potentially acquire or release, but it should not be the only tool used by teams when making decisions.

## 6.2 Limitations

There are various shortcomings and novelties of the data that are unable to be accounted for in the analysis. The most significant shortcoming is the fact that the data only contains statistics from the 2013-2014 and 2014-2015 NFL seasons, which manifests itself in the two ways described below.

An underlying assumption that is required to employ a 2 sample t-test is that the data from both populations are normally distributed. In practice, the Central Limit Theorem is often invoked, which states that the distribution of the sum of a sufficiently large sample of independent, identically distributed random variables will be approximately normal. This theorem is invoked when using the 2 sample t-tests in **Section 4.3**, but the size of some clusters may not be considered sufficiently large.

However, as the dataset grows and more seasons are added, it is expected that the majority of clusters will be sufficiently large to invoke the Central Limit Theorem, provided that the optimal number of clusters $k^*$ does not increase proportionally to the increase in size of the dataset.

Another concern with the small sample size involves the distribution fitting process as described in **Section 4.4**. When fitting a distribution to historical data, the best fit is determined mainly based on its goodness of fit relative to other candidate distributions, with the exception of the inspection of the P-P and Q-Q plots. As seen in **Appendix D**, the P-P and Q-Q plots suggest that the majority of distributions that were fit are relatively good fits. However, some clusters had too small of samples to fit an appropriate distribution and thus were discarded from the analysis. This is not to say that the players from these clusters are valued correctly, but there is not sufficient evidence to say that they are overvalued or undervalued.

## 6.3 Future Research

The analysis conducted in this work revolved primarily around using past performance to determine whether a player warranted the salary he was paid. While this is one way to approach the task of identifying overvalued and undervalued players, another approach is to try to map past performance to future performance based on other machine learning techniques classified as "semi-supervised" and "supervised" learning techniques (Chapelle et al., 2006). Using these techniques, the dataset would be used as training data, in which each observation would include an input object and a desired output value. The input object would be the input vector of characteristics, while the output value would be the player's salary in either the current season or season that follows. An inferred function would then be determined, which could be used to map new input vectors to a given salary, providing teams with an estimate of a player's salary given his characteristics for a given season. It is worth noting that similar limitations may arise regarding the small sample size or the ability of the algorithm to accurately map input vectors to salaries.

Another alternative to the methodology described above is to further explore the salary regression model described in **Section 4.1**. If an accurate explanatory or predictive salary model could be developed using regression techniques, it would be able to inform NFL executives of how certain characteristics are precisely valued in the labor market, as well as provide them with a model that can precisely determine the salary a player deserves based on his performance. The most difficult aspect of developing an accurate salary model is being able to control for the various factors that impact an offensive lineman's performance. This work attempts to create the control through the use of "differential statistics", as described in **Section 3**. Some limitations of the use of "differential statistics" are described in **Section 3**, and it is encouraged to use more descriptive data if enhanced player data tracking techniques are employed in the future by sports analytics companies.

Furthermore, it may be the case that the outcomes used to derive these statistics are a function of strategic decisions made by teams. For example, a defensive team may purposely place their best defensive lineman on the side of the line as the worst offensive lineman on the opposing team, thus exacerbating that offensive lineman's poor statistics. Furthermore, if the best offensive lineman is on the opposing side of the line in this situation, then his statistics are artificially improved since the lineman on the other side is playing so poorly. Thus, additional controls may be needed to develop an accurate explanatory of predictive salary model. Potential controls include the ability of the defensive lineman on the field and the in-season performance statistics of the defense that is being played.

## 7. Conclusion

Evaluating the performance of individual offensive linemen is a task that has been difficult to accomplish without the extensive use of rating players by watching game film. The most widely cited metric that is used to quantify the performance of offensive linemen is the statistic created by Pro Football Focus, which is calculated based off of individual grades given to offensive linemen based on an assessment of game film. This work aims to create an evaluative approach that is based off of objective statistics gathered relating to each player, with the goal of emphasizing the actual outcome of a play rather than subjectively assessing how the player's actions contributed to the outcome.

Through a multi-step methodology that groups similar players into clusters and subsequently evaluates the salary distribution of the clusters, certain players are identified as being overvalued or undervalued based on the salary that they were paid in the given year. This list of players can be used by NFL executives as a method to narrow down the list of potential players to target in the offseason. Using the dataset obtained for this paper, 3 players were found to be overvalued and 3 players were found to be undervalued in the 2013-2014 and 2014-2015 NFL seasons. The Pro Football Focus metric was then used as a proxy for player performance, and it was found that 4 of the 6 players significantly over performed or underperformed relative to their salary in the same direction as identified by the analysis. As more seasons worth of data become available, the hope is to conduct a more robust analysis that can be used by NFL executives across the league to make more informed decisions regarding the acquisition and release of offensive linemen in the National Football League.

## 8. Acknowledgements

We would like to acknowledge Ryan Warkins, Matt Scott and STATS, LLC for generously providing the dataset used in the analysis.

# Appendix A: Differential Statistics Methodology

To further investigate the efficacy of utilizing the "not to side" statistics in the differential statistics, as opposed to only using the "to side" statistics, regressions were run to determine the value of including the "not to side" statistics in the analysis. Using the methodology outlines in **Appendix B**, the stepwise regression was run using "to side" statistics instead of differential statistics. The results of this regression are presented in **Table 8**. The Adjusted R-squared from this regression is 0.47.

| Variable | Estimate | t value | Pr(>|t|) |
|---|---|---|---|
| (Intercept) | 2945845 | 2.22 | 2.87E-02 |
| Avg, PFF Rating Prior to Contract | 58801 | 3.143 | 0.002206 |
| Yards per Attempt to Side | 664269 | 2.215 | 0.029027 |
| Successful Run % to Side | -7186 | -2.872 | 4.99E-03 |
| Experience | -216627 | -2.918 | 0.004359 |
| Draft Round | -275462 | -4.333 | 0.0000353 |
| Pro Bowl Selections | 595882 | 3.457 | 0.000808 |

**Table 8:** Results of regression model with "To Side" statistics

To determine if "Not to Side" statistics are important in explaining a player's salary, the "Not to Side" statistics for "Yards per Attempt" and "Successful Run %" were added to the regression to examine the effect on the model. The results of this subsequent regression with an Adjusted R-squared of 0.5 are in **Table 9**.

| Variable | Estimate | t value | Pr(>|t|) |
|---|---|---|---|
| (Intercept) | 3823923 | 2.434 | 0.016753 |
| Avg, PFF Rating Prior to Contract | 57296 | 3.043 | 0.003014 |
| Yards per Attempt to Side | 802234 | 2.446 | 0.016235 |
| Yards per Attempt Not to Side | -369593 | -1.152 | 0.252264 |
| Successful Run % to Side | -39231 | -1.124 | 0.263812 |
| Successful Run % Not to Side | 33745 | 0.951 | 0.344178 |
| Experience | -204426 | -2.709 | 0.007981 |
| Draft Round | -276854 | -4.314 | 3.86E-05 |
| Pro Bowl Selections | 601169 | 3.475 | 0.000766 |

**Table 9:** Results of regression model including "Not to Side" statistics

As seen in **Table 9**, including the "Not to Side" statistics impacts the model in a non trivial manner. Both of the "Not to Side" statistics have a coefficient that is opposite in sign to the "To Side" statistic, implying that the "Not to Side" statistics have the opposite effect on a player's salary than the "To Side" statistics – exactly what would be expected if the "Not to Side" statistics served as an adequate control for the ability of other players on the field (e.g. running back, linemen on the other side).

Furthermore, including the "Successful Run % Not to Side" statistic causes the "Successful Run % To Side" statistic to no longer be significant in the regression. This is further evidence that "Not to Side" statistics are important to consider in the model. If the "Not to Side" statistics were not important, than they would not affect the significance of the "To Side" statistics, and would potentially have the same sign as well. Although not all variables are significant in the regression, the evidence discussed above suggests that incorporating "Not to Side" statistics is important to develop a stronger regression model. Based on this information, and the fact that the regression model with the "differential statistics" has a higher adjusted R-squared value, we believe that the use of "differential statistics" serves as a reasonable control given the limitations of the data.

# Appendix B. Labor Market Pricing of Offensive Lineman

This Appendix details the application of the methodology outlined in **Section 4.1** to the dataset used for the analysis.

The initial set of potential variables includes 15 variables corresponding to the data described in **Section 3.** This includes demographic characteristics (e.g. age, experience, draft round and pick), proxies for past performance (e.g. All Pro and Pro Bowl selections, past PFF +/- ratings) and 5 "differential statistics" for the year in question believed to be important in describing a lineman's performance (e.g. stuff percentage differential, rushing yards per carry, successful run percentage, pressures allowed percentage, sacks allowed percentage). The player's PFF rating for the year in question was also included as a potential predictor, to assess whether it is a stronger representation of the player's on field performance in a given year than the "differential statistics" used. **Table 10** provides a further description of the differential statistics used in the analysis.

| Differential Statistic | Description |
|---|---|
| Stuff % differential | % of carries to lineman's side minus % of carries not to lineman's side that were stopped at or behind the line of scrimmage |
| Yds/Attempt differential | Yards per carry to lineman's side minus yards per carry not to lineman's side |
| Successful Run % differential | % of carries to lineman's side minus % of carries not to lineman's side that resulted in a "successful" outcome |
| Pressures allowed % | Pressures allowed by lineman divided by pressures allowed by other linemen on field |
| Sack % | Sacks allowed by lineman divided by sacks allowed by other linemen on field |

Note: A successul play occurs on 1st down when a play achieves greater than 40% of the yardage to gain, on 3nd down when it achieves greater than 50% of yardage to gain, on 3rd or 4th down when play results in a 1st down. Touchdowns are always successful plays; turnovers are always unsuccessful plays

**Table 10:** Description of "differential statistics" used in regression analysis

The model specified by the stepwise regression included 8 independent variables, having an adjusted R-squared value of .50. **Table 11** shows the coefficients associated with the model, which will be referred to as "initial salary model."

Using the initial salary model, the Performance and Experience metrics were created for each player and the "Team Experience Metric" and "Team Performance Metric" was calculated for each player, as described in **Section 4.1**. The categorization of variables into performance and experience variables can be found in **Table 12**.

A new stepwise linear regression was then run that included these two metrics and their squared terms as independent variables. However, the final model determined by this stepwise selection process was deemed to fit the data worse than the initial salary model. Thus, the controls for the players on the line to both sides of a given player were not found to be significant and were not included in the final regression model. Instead, the initial salary model was used for the remainder of the analysis.

There are a few possible reasons that the controls were not found to be significant in the new regression model. A given player's salary at the time a contract is signed is a function of their past performance, as well as their future expected performance. The initial salary model attempts to control for the future expected performance of players on the other side of the line (as well as the running back and other players on the field) through the use of differential statistics, using actual performance as a proxy. The "Team Performance" metric attempts to control for the future expected performance of players on the same side of the line. However, players on the same side of the line may have similar enough statistics from the current season to the lineman in question that it does not provide additional explanatory power in a regression model. The "Team Experience" metric explores whether the other linemen's past performance has an impact on the given lineman's salary. In theory, an NFL team may try to sign players who they believe to be worth less money if their other linemen are strong and the team has stronger needs at other positions. Based on this, the past performance of the other linemen may help explain the salary of a given lineman. However, based on the regression, this is not the case. It is not surprising that the "Team Experience" metric was not found to be significant, but we wanted to explore the possibility to conduct as thorough of an analysis as possible.

| Variable | Estimate | t value | Pr(>|t|) |
|---|---:|---:|---:|
| (Intercept) | 5399261 | 9.427 | 2.12E-15 |
| Avg, PFF Rating Prior to Contract | 56697 | 3.081 | 0.00268 |
| Experience | -199134 | -2.694 | 0.00831 |
| Draft Round | -264405 | -4.224 | 5.38E-05 |
| Pro Bowl Selections | 624910 | 3.715 | 0.000338 |
| Stuff % Differential | -82247 | -2.551 | 0.012284 |
| Yds per Attempt Differential | 382197 | 2.035 | 0.044516 |
| Sack % | -2516 | -2.31 | 0.022994 |

**Table 11:** Initial salary model regression results

| Experience | Performance |
|:---:|:---:|
| Age | Stuff % differential |
| Experience (Years in League) | Yds/Attempt differential |
| Draft_Pick | YAC/Attempt Differential |
| 1st Team All Pro | YBC/Attempt Differential |
| 2nd Team All Pro | Successful Run % Differential |
| PWF All Pro | Pressure Allowed Differential |
| Pro Bowl | Pressure % |
| PFF rating from past seasons | Sack % |
|  | Rush Tds/Attempts Differential |
|  | Att/Dropback |
|  | PFF rating from curent season |

**Table 12**: Categorization of independent variables into experience vs performance variables

# Appendix C. Clustering Analysis

This Appendix details the application of the methodology outlined in **Section 3.2** to the dataset used for the analysis.

Using the characteristics from the "Salary Model," input vectors were created for each of the players in the dataset as outlined in **Section 4.2**. The input vectors were then used to run the k-means clustering algorithm for $k \in \{1, \ldots, 20\}$. The Krzanowski-Lai statistic was then calculated for each value of $k$, with the results shown in **Figure 1.**

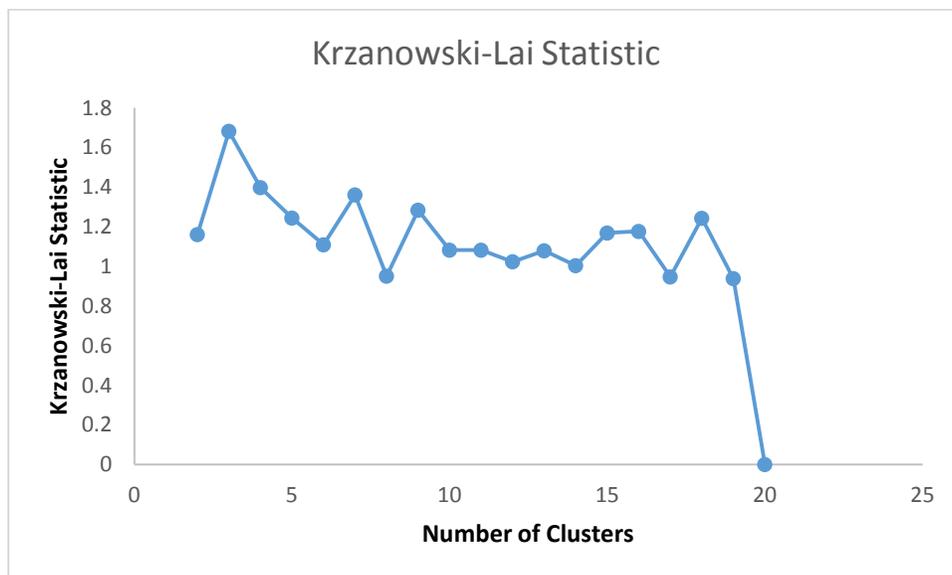

**Figure 1:** Krzanowski-Lai statistic for various values of $k$

Based on **Figure 1**, it was determined that $k^* = 7$ since there exists a local maxima at $k = 7$ and the marginal decrease in the Krzanowski-Lai statistic from $k = 7$ to subsequent local maxima is not as large. Thus, the k-means algorithm was run with 7 clusters to group players together for the remainder of the analysis. The resulting cluster that each player was placed in can be found in **Table 13**. It is worth noting that if a player that is listed twice in **Table 13**, then the first entry corresponds to the player's performance in 2013 and the second entry corresponds to the player's performance in 2014.

| Player | Cluster | Player | Cluster | Player | Cluster |
|---|---|---|---|---|---|
| Alex Boone | 4 | Evan Mathis | 2 | Mike Brisiel | 5 |
| Andre Smith | 1 | Evan Mathis | 2 | Mike McGlynn | 3 |
| Andrew Gardner | 4 | Fernando Velasco | 5 | Mike McGlynn | 3 |
| Andy Levitre | 1 | Gosder Cherilus | 1 | Nate Chandler | 4 |
| Andy Levitre | 3 | Gosder Cherilus | 1 | Nick Hardwick | 2 |
| Anthony Collins | 3 | Harvey Dahl | 5 | Paul Fanaika | 4 |
| Austin Howard | 4 | J.D. Walton | 7 | Paul McQuistan | 3 |
| Austin Howard | 5 | Jake Long | 6 | Phil Loadholt | 1 |
| Ben Grubbs | 1 | Jared Veldheer | 3 | Phil Loadholt | 3 |
| Ben Grubbs | 7 | Jeremy Zuttah | 7 | Ramon Foster | 4 |
| Brad Meester | 2 | Jermon Bushrod | 1 | Ramon Foster | 4 |
| Branden Albert | 1 | Jermon Bushrod | 1 | Roberto Garza | 2 |
| Branden Albert | 1 | Jeromey Clary | 5 | Roberto Garza | 2 |
| Breno Giacomini | 4 | Joe Barksdale | 1 | Rodger Saffold | 3 |
| Breno Giacomini | 4 | Joe Barksdale | 7 | Ryan Clady | 6 |
| Brian de la Puente | 4 | Joe Berger | 2 | Ryan Kalil | 6 |
| Brian de la Puente | 5 | John Jerry | 3 | Ryan Kalil | 6 |
| Bryant McKinnie | 1 | Jon Asamoah | 1 | Ryan Wendell | 4 |
| Byron Bell | 4 | Jonathan Goodwin | 2 | Ryan Wendell | 5 |
| Chad Rinehart | 3 | Josh Sitton | 2 | Samson Satele | 3 |
| Chad Rinehart | 7 | Justin Blalock | 3 | Samson Satele | 7 |
| Charlie Johnson | 5 | Justin Blalock | 7 | Scott Wells | 5 |
| Charlie Johnson | 5 | Kevin Boothe | 5 | Scott Wells | 5 |
| Chris Chester | 1 | Khalif Barnes | 1 | Sebastian Vollmer | 1 |
| Chris Chester | 1 | Khalif Barnes | 1 | Shawn Lauvao | 3 |
| Chris Clark | 4 | King Dunlap | 4 | T.J. Lang | 3 |
| Chris Myers | 2 | King Dunlap | 4 | T.J. Lang | 3 |
| Chris Myers | 2 | Kory Lichtensteiger | 7 | Ted Larsen | 4 |
| Chris Williams | 7 | Kory Lichtensteiger | 7 | Todd Herremans | 2 |
| Dan Connolly | 5 | Kraig Urbik | 7 | Todd Herremans | 2 |
| Dan Connolly | 5 | Kraig Urbik | 7 | Tony Pashos | 2 |
| Daryn Colledge | 3 | Kyle Cook | 5 | Travelle Wharton | 2 |
| Daryn Colledge | 3 | Logan Mankins | 6 | Tyler Polumbus | 4 |
| Davin Joseph | 7 | Logan Mankins | 6 | Tyson Clabo | 4 |
| Davin Joseph | 7 | Louis Vasquez | 1 | Will Beatty | 1 |
| Donald Penn | 4 | Louis Vasquez | 7 | Will Beatty | 1 |
| Doug Free | 1 | Lyle Sendlein | 5 | Will Montgomery | 5 |
| Doug Free | 1 | Lyle Sendlein | 5 | Will Montgomery | 5 |
| Doug Legursky | 5 | Mackenzy Bernadeau | 5 | Willie Colon | 2 |
| Eric Winston | 1 | Manny Ramirez | 7 | Willie Colon | 2 |
| Erik Pears | 4 | Manny Ramirez | 7 | Zach Strief | 4 |
| Erik Pears | 5 | Marshal Yanda | 6 | Zach Strief | 5 |
| Eugene Monroe | 1 | Marshal Yanda | 6 | Zane Beadles | 7 |
| Evan Dietrich-Smith | 4 | Matt Slauson | 5 | | |
| Evan Dietrich-Smith | 4 | Michael Oher | 3 | | |

**Table 13:** Cluster placement of players in dataset

# Appendix D. Distribution Fitting

This Appendix details the application of the methodology outlined in **Section 4.4** to the dataset used for the analysis.

Following the characterization of clusters, distributions were fit to the salary data of each of the 6 clusters that were deemed to be of large enough size. The most appropriate fit was determined based on the criteria outlined in **Section 4.4**. A sample distribution fit, and the corresponding P-P and Q-Q plots can be found for Cluster 1 in **Figures 2, 3, and 4**.

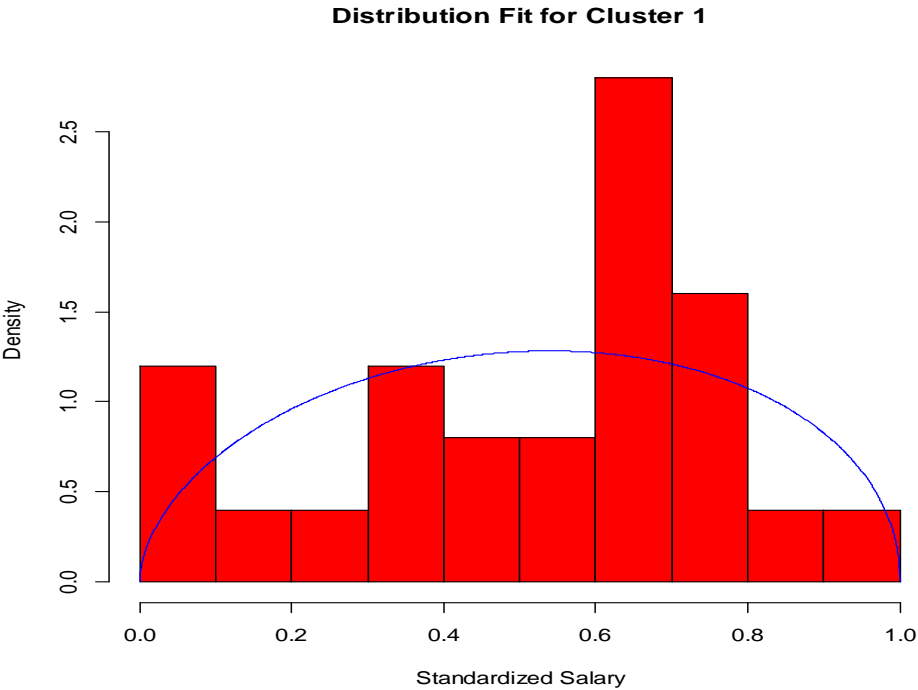

**Figure 2**: Beta distribution fit for Cluster 1

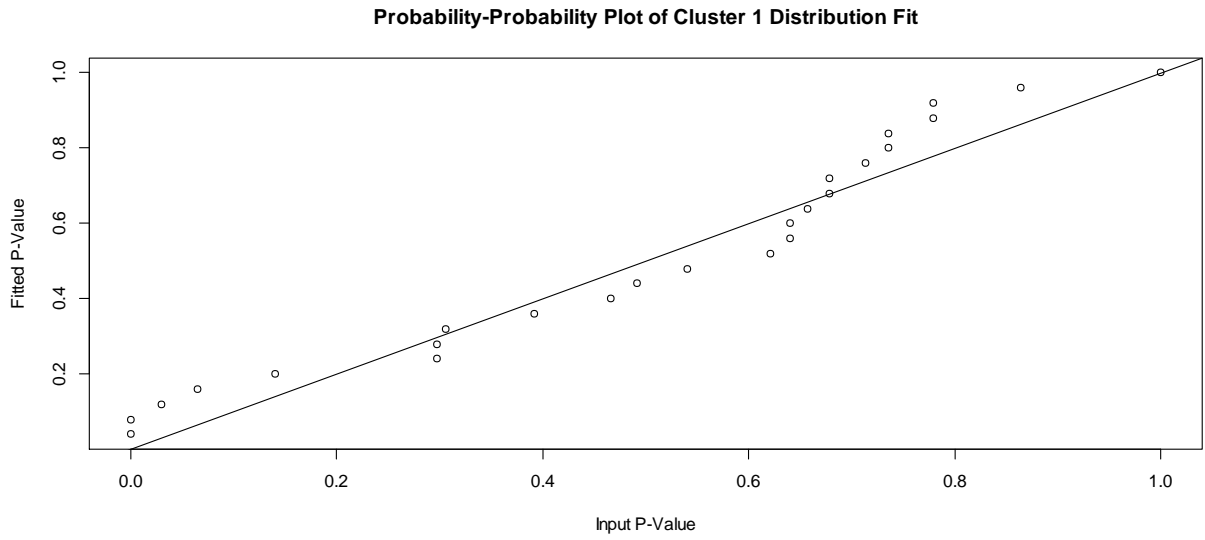

**Figure 3:** Probability-Probability (P-P) plot for the beta distribution fit of Cluster 1

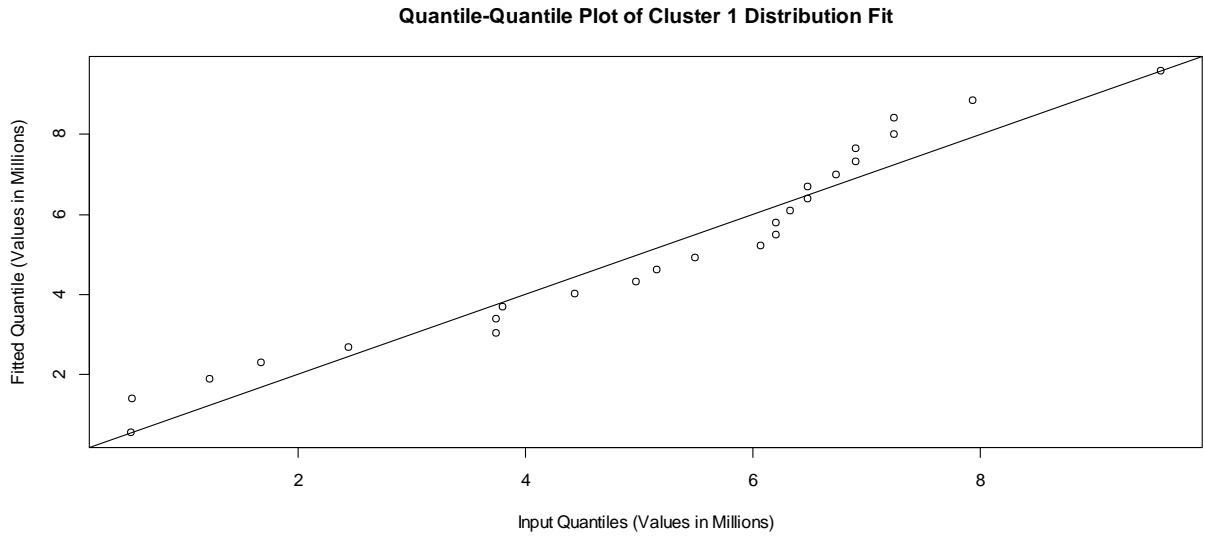

**Figure 4:** Quantile-Quantile (Q-Q) plot for the beta distribution fit of Cluster 1